\pdfoutput=1
\documentclass[11pt]{article}

\usepackage[final]{acl}

\usepackage{times}
\usepackage{latexsym}

\usepackage[T1]{fontenc}

\usepackage[utf8]{inputenc}

\usepackage{microtype}

\usepackage{inconsolata}

\usepackage{graphicx}
\usepackage[utf8]{inputenc} 
\usepackage[T1]{fontenc}    
\usepackage{hyperref}       
\usepackage{url}            
\usepackage{booktabs}       
\usepackage{amsfonts}       
\usepackage{nicefrac}       
\usepackage{microtype}      
\usepackage{xcolor}         
\usepackage{marvosym}
\usepackage{multirow}
\usepackage{makecell}
\usepackage{colortbl}
\usepackage{tcolorbox}
\usepackage{longtable}
\usepackage{subcaption}
\usepackage{pifont}
\usepackage{framed}
\usepackage[utf8]{inputenc}
\usepackage{amsmath}
\usepackage{enumitem}
\usepackage{color}
\usepackage{array}
\usepackage{diagbox}
\usepackage{arydshln}

\usepackage{tikz}
\usepackage[edges]{forest}
\usepackage{pgfplots}
\usepackage{bm}
\usepackage{scalefnt}
\usepackage{subcaption}
\definecolor{lgreen}{rgb}{0.89,0.94,0.85}
\definecolor{lred}{rgb}{0.98, 0.90, 0.84}
\definecolor{lyellow}{rgb}{1.00, 0.95, 0.80}
\definecolor{lblue}{rgb}{0.85, 0.89, 0.95}
\definecolor{hidden-draw}{RGB}{20,68,106}
\definecolor{hidden-pink}{RGB}{255,245,247}

\usepackage{geometry}
\usepackage{tabularx}
\usepackage{booktabs}
\usepackage{enumitem}

\title{IOPO: Empowering LLMs with Complex Instruction Following via Input-Output Preference Optimization}

\author{%
  Xinghua Zhang, 
  Haiyang Yu,
  Cheng Fu,
  Fei Huang,
  Yongbin Li\thanks{Corresponding author.}\\
  Tongyi Lab, Alibaba Group \\
  \normalsize{
\{\texttt{zhangxinghua.zxh}, \texttt{yifei.yhy}, \texttt{fucheng.fuc}, \texttt{f.huang}, \texttt{shuide.lyb}\}@alibaba-inc.com}
}

\begin{document}
\maketitle

\begin{abstract}
In the realm of large language models (LLMs), the ability of models to accurately follow instructions is paramount as more agents and applications leverage LLMs for construction, where the complexity of instructions are rapidly increasing. 
However, on the one hand, there is only a certain amount of complex instruction evaluation data; on the other hand, there are no dedicated algorithms to improve the ability to follow complex instructions. 
To this end, this paper introduces \textbf{\textsc{Trace}}, a benchmark for improving and evaluating the complex instruction-following ability, which consists of 120K training data and 1K evaluation data.
Furthermore, we propose \textbf{\textsc{IOPO}} (Input-Output Preference Optimization) alignment method which takes both input and output preference pairs into consideration, where LLMs not only rapidly align with response preferences but also meticulously explore the  instruction preferences.
Extensive experiments on both in-domain and out-of-domain datasets confirm the effectiveness of \textsc{IOPO}, showing 8.15\%, 2.18\% improvements on in-domain data and 5.91\%, 2.83\% on out-of-domain data compared to SFT and DPO respectively.
Our code and dataset are released at \url{https://github.com/AlibabaResearch/DAMO-ConvAI/tree/main/IOPO}.
\end{abstract}
\section{Introduction}
The rapid development of LLMs has facilitated human-machine interaction, with instructions serving as the medium~\cite{gao2024taxonomy,kim2024understanding,zhang2024imperative,wang2024reframing,zhang2024revealing}. As human needs evolve, there is an increasing expectation for models to handle more intricate tasks through complex instructions~\cite{ge2023openagi,yang2024harnessing,wang2024leave}. Consequently, the instruction-following ability, especially complex instructions, is garnering significant attention~\cite{zhou2023instruction,xu2024wizardlm,li-etal-2024-cif,zhang2024cfbench}.

\begin{figure}[t]
\centering
\includegraphics[width=1.0\columnwidth]{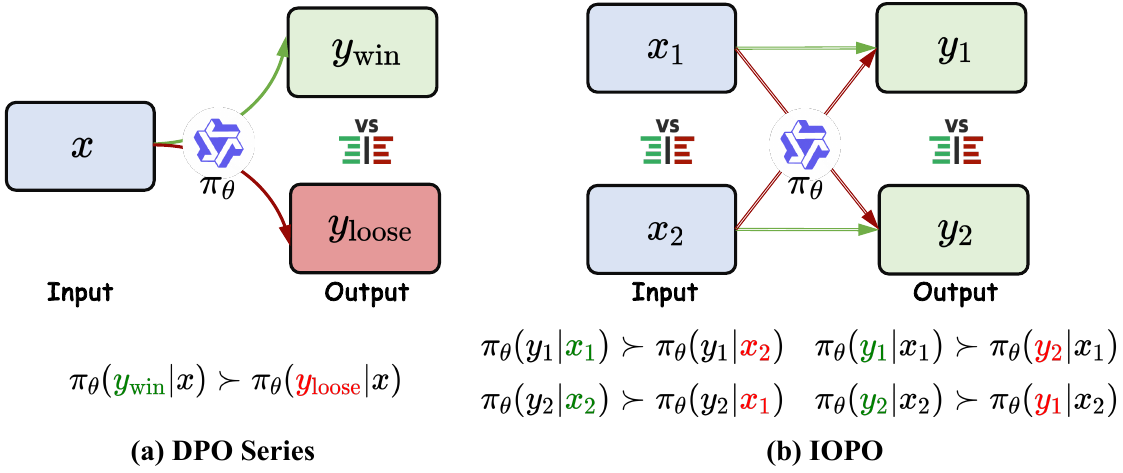} 
\caption{Alignment Paradigms (a) Existing DPO Series vs. (b) Proposed IOPO. The green arrow indicates that $y$ matches $x$ while the red one indicates a mismatch.}
\label{fig_cmp}
\end{figure}

To evaluate the instruction-following abilities of LLMs, several benchmarks~\cite{zhou2023instruction,qin-etal-2024-infobench,li-etal-2024-cif} have been proposed, which are designed to systematically assess how well these models can understand and execute instructions. 
IFEval~\cite{zhou2023instruction} focuses on verifiable instructions which are amenable to objective verification of compliance.
\citet{qin-etal-2024-infobench} introduces \textsc{InFoBench} which contains 2,250 decomposed questions to assess the instruction following.
Recently, the ability to follow complex instructions with multiple constraints is gaining increasing attention~\cite{he2024can,jiang-etal-2024-followbench,wen2024benchmarking,he2024complex} as LLMs are deployed in sophisticated real-world applications.
\citet{zhang2024cfbench} proposes a constraint-following benchmark CFBench with 1,000 multi-constraint samples.
However, most of benchmarks lay emphasis on evaluating LLMs' ability to follow complex instructions, lack of algorithms tailored for enhancing the corresponding ability.

From RLHF (Reinforcement Learning from Human Feedback)~\cite{Long-etal-training,bai2022training} to the following-up researches such as DPO (Direct Preference Optimization)~\cite{Rafael-direct-2023-etal}, alignment algorithms which align LLMs with human preferences, have demonstrated their effectiveness in improving the LLMs' capabilities to follow instructions.
Nevertheless, these methods directly explore different responses $y$ ($y_{\rm win}$, $y_{\rm loose}$) based on the same instruction $x$, as shown in Figure~\ref{fig_cmp} (a).
In the complex instruction scenario which contains multiple constraints, it is challenging to efficiently perceive the fine-grained constraints in $x$ solely by modeling different $y$.

To bridge this gap, this paper first introduces \textbf{\textsc{Trace}} benchmark to improve the ability of LLMs to \textbf{t}rack complex fine-g\textbf{r}ained constr\textbf{a}int instru\textbf{c}tions and make them more ob\textbf{e}dient. 
\textsc{Trace} is automatically constructed based on the manually sorted taxonomy of complex instructions with 26 constraint dimensions within 5 constraint types.
We develop an automated data construction workflow that extends from open-source simple instructions to multi-constrained complex ones.
In the end, we accumulate 120K complex instructions for model training and 1K human-verified data for evaluation.
To enhance the ability of LLMs to follow complex instructions, this paper further proposes {\it Input-Output Preference Optimization} (\textbf{\textsc{IOPO}}) method.
IOPO not only takes the instruction $x$ as input to directly learn the response $y$ preference, but also gradually delves deeper into instructions $x$ based on the same response $y$, to promote effective perception of fine-grained constraints, as shown in Figure~\ref{fig_cmp} (b).
The major contributions of this paper are summarized as follows:
\begin{itemize}
\item We introduce a benchmark \textbf{\textsc{Trace}} for complex instruction following, which includes both an evaluation set and a training set, and an automated data construction workflow, further enriching the research community.

\item Different from previous alignment paradigm, we propose \textbf{\textsc{IOPO}} alignment method which deeply explores the complex instructions $x$ ({\it Input}), not just directly learning response preference $y$ ({\it Output}).

\item Extensive experiments on both in-domain and out-of-domain evaluations have confirmed the consistent improvements, with an average increase of 7.03\% and 2.51\%, compared to SFT and DPO, respectively.

\end{itemize}

\section{Related Work}
\subsection{Instruction Following}
Instruction following is the most fundamental and crucial ability for large language models (LLMs), which enables them to understand and execute user instructions accurately, making them more effective in a wide range of applications.
In fact, earlier studies have explored the extent to which models follow language instructions~\cite{ye-ren-2021-learning,mishra-etal-2022-cross,hase-bansal-2022-models}.
It is effective to fine-tune LLMs on these annotated instruction data for improving the ability to follow natural language instructions.
The instruction-following ability enhances adaptability to unseen tasks, which has become an efficient learning paradigm for novel task demands~\cite{lou2023comprehensive}.

As human demands grow higher, the instructions given to LLMs are also becoming increasingly complex.
Recent studies are beginning to focus on the complex instruction-following ability of LLMs, where more complex or constrained instructions have been proven effective in enhancing LLMs' abilities to follow instructions~\cite{mukherjee2023orca,xu2024wizardlm,luo2024wizardcoder}.
Constrained instructions, as a type of complex instruction, are also gradually receiving attention from the research community.
Increasing the complexity of constraints within the instruction (e.g., raising the number of constraints) can further improve the ability to follow complex instructions~\cite{sun2024conifer,dong2024self,he2024complex}.
\citet{sun2024conifer} introduces a instruction tuning dataset Conifer and proposes a progressive learning scheme to enhance the ability of LLMs to follow multi-level instructions with complex constraints.
\citet{he2024complex} first finds that multi-constraint instructions can enhance LLMs' understanding of complex instructions, and then introduces a discrimination-based method for data acquisition, and finally proposes a contrastive method with reinforcement learning fine-tuning (RLFT) for data utilization.
In addition, some work focuses on evaluating the multi-constraint instruction-following capabilities of LLMs~\cite{zhou2023instruction,qin-etal-2024-infobench,jiang-etal-2024-followbench}.
\citet{zhang2024cfbench} introduces CFBench, a benchmark which encompasses instructions with multiple constraints, and proposes a multi-dimensional evaluation framework to comprehensively assess model capabilities.

\subsection{LLM Alignment}
LLM alignment aims to enhance LLMs by aligning them with human preference.
Recent research has conducted extensive explorations for LLM alignment.
From RLHF/PPO~\cite{Long-etal-training,bai2022training} to DPO~\cite{Rafael-direct-2023-etal} and beyond~\cite{bai2022constitutional,song2024preference,meng2024simpo}, the evolution of xPO-series alignment algorithms has seen significant advancements. These methods have been pivotal in improving the alignment between LLMs and human values, ensuring that the outputs of these models are not only effective but also follow human preference.

RLHF involves training models using human-provided rewards or reward models to improve decision-making, optimizing policies through iterative feedback loops for more aligned outcomes.
DPO directly optimizes the model's output to match preferred response as indicated by human feedback, which simplifies the alignment process by focusing on direct comparisons between preferred and dispreferred outputs, allowing the model to learn human preference without needing an explicitly defined reward model.
SimPO~\cite{meng2024simpo} proposes a method for preference optimization that eliminates the need for the reference model $\pi_{\rm ref}$, which is memory efficient and simple.
Instead of using pairwise data, PRO~\cite{song2024preference} utilizes the preference ranking of any length listwise preference dataset.
ORPO~\cite{hong-etal-2024-orpo} introduces the reference model-free preference optimization algorithm.

To further enrich the community of multi-constraint instruction following ability, we construct \textsc{Trace} benchmark which contains instructions with multiple constraints, more fine-grained constraint types and a wider range of constraint quantities.
In addition, we propose the tailor-designed alignment algorithm IOPO for fine-grained multi-constraint alignment, which is different from previous methods (e.g., DPO) only focusing on the output preference.
\section{Preliminary}
Existing alignment methods have evolved from RLHF~\cite{Long-etal-training,bai2022training}, this section provides a brief introduction to RLHF which mainly consists of three stages:

1) \textbf{SFT}: The generic pre-trained LM is fine-tuned with maximum likelihood supervised loss on downstream task data, and then we can get the SFT model $\pi_{\rm SFT}$.

2) \textbf{Reward Model}: The model $\pi_{\rm SFT}$ is utilized with prompt $x$ to generate two different responses $y_1$, $y_2$. The pair of responses is labeled as ``preferred'' and ``dispreferred'' by human labelers, i.e., $y_1 \succ y_2$ | $x$. The reward model $r_{\phi}$ is trained with the following negative log-likelihood loss:
\begin{align}
    \mathcal{L}_{R} = -\mathbb{E}_{(x, y_{1}, y_{2}) \sim \mathcal{D}}[\log \sigma(r_{\phi}(x, y_{1}) - r_{\phi}(x, y_{2}))]
\label{rm-eq}
\end{align}

3) \textbf{RL}: This stage uses the learned reward model $r_{\phi}$ to provide feedback to the language model policy, the optimization objective is as follows:
\begin{equation}
\begin{aligned}
\max_{\pi_\theta} \mathbb{E}_{x \sim D, y \sim \pi_\theta(y | x)}[r_\phi(x, y)] - \\ \beta \mathbb{D}_\text{KL}[\pi_\theta(y | x) || \pi_\text{ref}(y | x)]
\label{rl-eq}
\end{aligned}
\end{equation}
where LM policy $\pi_{\theta}$, base reference policy $\pi_{\rm ref}$ are both initialized with $\pi_{\rm SFT}$, $\beta$ controls the deviation of $\pi_{\theta}$ from the base policy $\pi_{\rm ref}$.

\begin{figure*}[!t]
\centering
\includegraphics[width=1.9\columnwidth]{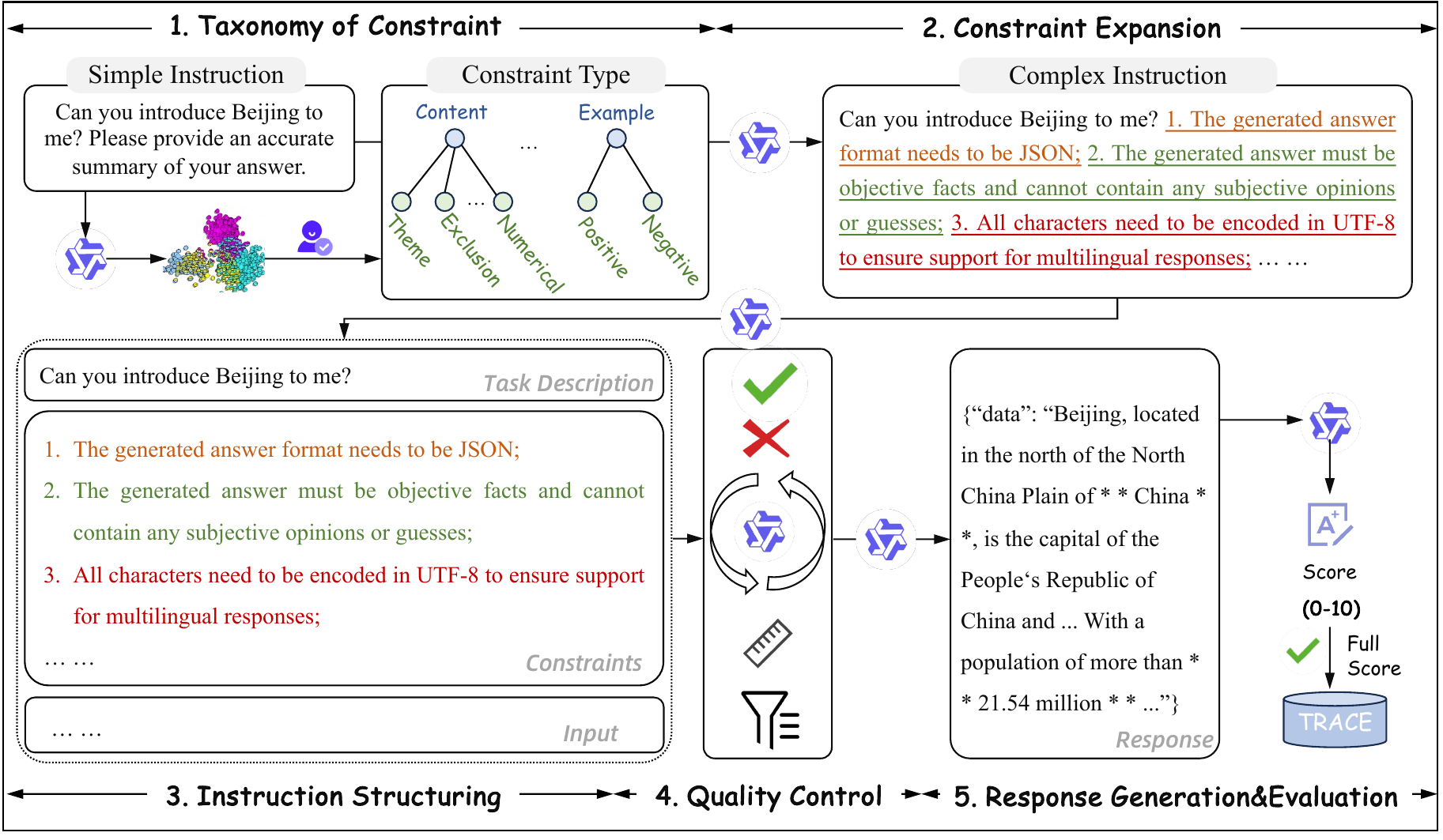} 
\caption{Construction Pipeline of \textsc{Trace}, which consists of five key stages: 1) {\it Taxonomy of Constraint}, 2) {\it Constraint Expansion}, 3) {\it Instruction Structuring}, 4) {\it Quality Control}, and 5) {\it Response Generation \& Evaluation}.}
\label{fig_data_ann}
\end{figure*}

\section{\textsc{Trace} Benchmark}
This section describes the construction pipeline of \textsc{Trace}, its statistics, and evaluation protocol.

\subsection{Construction Pipeline}
\label{trace-cp}
The overall construction process as shown in Figure~\ref{fig_data_ann} includes several key stages: 1) {\it Taxonomy of Constraint}, 2) {\it Constraint Expansion}, 3) {\it Instruction Structuring}, 4) {\it Quality Control}, and 5) {\it Response Generation \& Evaluation}.

\textbf{Taxonomy of Constraint.}
A comprehensive constraint type system is developed through inference by LLM from a large volume of open-source simple instructions, and further refined by human experts, into 5 constraint types~\cite{jiang-etal-2024-followbench} and 26 constraint dimensions. The detailed description of constraints is shown in Appendix~\ref{appendix:tax_constraint}.

\textbf{Constraint Expansion.}
This step aims to expand simple instructions into more complex ones that incorporate multiple constraints based on the taxonomy of constraint by prompting LLM.

\textbf{Instruction Structuring.}
To better distinguish different segments of the instruction, this step structures the flat instruction text expanded from the last step into {\it Task Description}, {\it Constraints}, and {\it Input} part by prompting LLM.

\textbf{Quality Control.}
To ensure the validity of the instructions, this step conducts quality control of the expanded instructions by prompting LLM, addressing some forms of invalidity such as redundancy between the description and constraints, incompleteness between the description and input.

\textbf{Response Generation \& Evaluation.}
First, we prompt LLM with the instruction $x$ to generate the corresponding response $y$.
To confirm its quality, we then prompt LLM to rate how well the response $y$ comply with constraints in the instruction $x$.
Finally, data that fully follows all the constraints outlined in the instruction would receive a perfect score of 10, and is selected to form the supervised fine-tuning (SFT) instruction dataset.

The corresponding prompts for constraint expansion, instruction structuring, quality control, response generation and evaluation are shown in Appendix~\ref{appendix:prompt}.

\begin{table}[t]
\renewcommand\arraystretch{1.1}
\tabcolsep=0.3cm
\centering
\small
\begin{tabular}{ccccc}
\toprule
& \#N & Min. & Max. & Avg.\\ \midrule
\#Training & 119,345  & 1  & 15 & 4.36  \\ 
\#Evaluation  & 1,042  & 1  & 15 & 4.89 \\ 
\bottomrule
\end{tabular}
\caption{The statistics of \textsc{Trace} benchmark. \#N is the number of instructions; Min., Max., and Avg. mean the minimum, maximum, and average number of constraints per instruction.}
\label{dataset-table}
\end{table}

\begin{table*}[t]
\renewcommand\arraystretch{1.1}
\tabcolsep=0.18cm
\centering
\small
\begin{tabular}{cccccccccccccccc}
\toprule
 $\mathcal{C}=$  & 1 &2 & 3 & 4 &5 & 6 & 7 &8 & 9 & 10 &11 & 12 & 13 &14 &15 \\ \midrule
 \#Training    & 991 & 8,003 & 26,421 & 34,155 & 26,327 & 13,858 & 5,882 & 2,185 & 999 & 464 & 8 & 20 & 20 & 8 & 4 \\ 
 \#Evaluation  &200 & 100 & 100 & 100 & 100 & 100 & 100 & 100 & 100 & 10 & 10 & 10 & 4 & 4 & 4\\
 \bottomrule
\end{tabular}
\caption{Constraint number ($\mathcal{C}$) distributions over training and evaluation set in \textsc{Trace}. $\mathcal{C}=i$ represents the number of instructions with $i$ constraints.}
\label{tb-cnum}
\end{table*}

\begin{figure}[t]
\centering
\includegraphics[width=0.9\columnwidth]{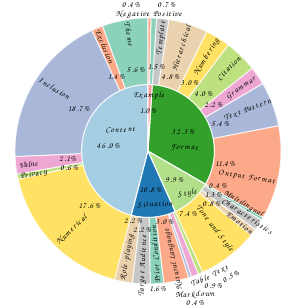} 
\caption{Constraint type distribution over evaluation set in \textsc{Trace}.}
\label{figctype}
\end{figure}

\subsection{Dataset Statistics}
As shown in Table~\ref{dataset-table}, \textsc{Trace} consists of 119,345 instructions for model training, and 1,042 instructions for evaluation, where the minimum and maximum number of constraints per instruction are 1 and 15, with average numbers of 4.36 and 4.89, respectively.
Table~\ref{tb-cnum} gives the constraint number distributions over training and evaluation set in \textsc{Trace}. For example, when $\mathcal{C}=6$, the corresponding column indicates that there are 13,858 instructions with 6 constraints in the training set, and 100 instructions with 6 constraints in the evaluation set.
In Figure~\ref{figctype}, we depict the constraint type distribution over the evaluation set. The inner circle represents the distribution of five major types ({\it Content Constraint}, {\it Situation Constraint}, {\it Style Constraint}, {\it Format Constraint}, and {\it Example Constraint}), and the corresponding outer one represents the distribution of concrete constraint dimensions.

\subsection{Evaluation Protocol} \label{eval_proto}
Following previous work~\cite{qin-etal-2024-infobench,zhang2024cfbench}, we use GPT-4o as the evaluator to assess the generated response based on the complex instruction.
Concretely, we prompt the LLM evaluator to evaluate each constraint mentioned in the complex instruction on a scale of 0--10, assessing the degree to which the response follows each constraint. A higher score indicates stronger adherence to the specified constraint. The overall instruction following score IF on the evaluation set with $n$ complex instructions are calculated as follows:
\begin{align}
   {\rm IF}=\frac{1}{n}\sum_{i}^{n} \mathbb{I}_{=10} \sum_{j=1}^{m_i} \frac{\mathcal{S}_{i,j}}{m_i}
\label{eval-eq}
\end{align}
where $\mathcal{S}_{i,j}\in[0,10]$ is the score indicating the degree to which the $j$-th constraint in the $i$-th instruction is adhered to. $m_i$ is the number of constraints in the $i$-th instruction. $\mathbb{I}_{=10}$ is 1 when $\sum_{j=1}^{m_i} \frac{\mathcal{S}_{i,j}}{m_i}=10$, otherwise is 0.
That is, a response is considered correct only when all constraints in the complex instruction are fully followed.

\subsection{Evaluation Set Quality}
To generate the high-quality evaluation set, we further introduce a rigorous post-inspection process after the construction pipeline (Sec.~\ref{trace-cp}).
First, based on LLM-as-Judge~\cite{wang-etal-2024-large-language-models-fair,zhang2023widerdeeperllmnetworks,zeng2024evaluating,xia-etal-2024-language}, we use the powerful LLM GPT-4o to check the following items for each instruction in the evaluation set:
1) Is the description empty?
2) Is there redundancy between the constraints and description?
3) Does the input match the description?
If any of the aforementioned issues arise, we prompt the GPT-4o to make corrections for accelerating the subsequent manual verification.
Second, the professional data annotation team makes the manual annotation process involving multiple steps such as annotator training, small-scale trial annotation, selection of official annotators, and formal annotation.
Finally, we randomly select 100 instructions for quality evaluation, which are then inspected by three labeling specialists based on the above check items and the overall validity. The agreement rate among three annotators on the sampled evaluation set is 95\%.

\section{Input-Output Preference Optimization}
Both RLHF~\cite{Long-etal-training,bai2022training} and its variants, such as DPO~\cite{Rafael-direct-2023-etal}, directly learn the response $y$ preference ({\it Output}) given the same instruction $x$ ({\it Input}).
However, complex instructions consist of multiple fine-grained constraints, direct preference learning for the output $y$ struggles to perceive fine-grained constraints in the input $x$.
To enhance the model's perception of fine-grained instruction, we further introduce the input preference learning which reflects on the constraints in the instruction $x$ based on the response $y$. 
By performing preference learning of both input and output, \textbf{i}nput-\textbf{o}utput \textbf{p}reference \textbf{o}ptimization (IOPO) not only rapidly fits the better output but also meticulously considers the fine-grained information of the input.

Concretely, we construct a pair of instructions <$x_1$, $x_2$> whose responses are respectively <$y_1$, $y_2$>, where $x_2$ has subtle differences from $x_1$ in some constraints, and these differences would result in substantially divergent responses.
And then, we can get four input-output pairs <$x_1$, $y_1$>, <$x_1$, $y_2$>, <$x_2$, $y_1$>, and <$x_2$, $y_2$>, which can form a preference group pair $\mathcal{G}_1$ $\succ$ $\mathcal{G}_2$ ($\mathcal{G}_1$ = \{<$x_1$, $y_1$>, <$x_2$, $y_2$>\}, $\mathcal{G}_2$ = \{<$x_1$, $y_2$>, <$x_2$, $y_1$>\}). 
The detailed data construction process is described in Appendix~\ref{iopo-data}.
The first group is the matched input-output pair while the second one is mismatched.
As derived in \citet{Rafael-direct-2023-etal}, the reward function $r(x, y)$ can be represented by the policy model $\pi_r$ as follows:
\begin{align}
\small
   r(x, y) = \beta \log \frac{\pi_r(y|x)}{\pi_{\text{ref}}(y|x)} + \beta \log Z(x)
\label{reward-eq}
\end{align}
where $Z(x) = \sum_y \pi_\text{ref}(y | x) \exp \left( \frac{1}{\beta} r(x, y) \right)$.

The Bradley–Terry model~\cite{bradley1952rank} is a probability model for the outcome of pairwise comparisons between items, groups, or objects.
Given a pair of items $i$ and $j$, it estimates the probability that the pairwise comparison $i$ $\succ$ $j$ turns out true~\cite{David2004mm}, as
\begin{align}
\small
p(i \succ j) = {p_i}/{(p_i + p_j)}
\label{bt-eq}
\end{align}
where $p_i$ is a positive real-valued score assigned to individual $i$, and the comparison $i$ $\succ$ $j$ can mean ``$i$ is preferred to $j$''.
Similarly, given a pair of groups $\mathcal{G}_1$ and $\mathcal{G}_2$, we can define $p_1 = e^{r(x_1, y_1)+r(x_2, y_2)}$ for $\mathcal{G}_1$, and $p_2 = e^{r(x_1, y_2)+r(x_2, y_1)}$ for $\mathcal{G}_2$, as
\begin{equation}
\begin{aligned}
\small
p(\mathcal{G}_1 \succ \mathcal{G}_2) = \frac{e^{r_{\mathcal{G}_1}}}{e^{r_{\mathcal{G}_1}} + e^{r_{\mathcal{G}_2}}} \\
r_{\mathcal{G}_1} = r(x_1, y_1)+r(x_2, y_2)\\
r_{\mathcal{G}_2} = r(x_1, y_2)+r(x_2, y_1)
\label{bt-adapt-eq}
\end{aligned}
\end{equation}
Next, combining Eq.~\ref{bt-adapt-eq} and Eq.~\ref{reward-eq}, we can further derive as follows (details in Appendix~\ref{appendix:derivation}):
\begin{equation}
\small
\begin{aligned}
p(\mathcal{G}_1 \succ \mathcal{G}_2) = \sigma\left(\frac{1}{2}(\Pi_1+\Pi_2)\right)
\end{aligned}
\label{bt-adapt-f-eq}
\end{equation}
\begin{equation}
\small
\begin{aligned}
\Pi_1 = &2\beta\log\frac{\pi_r(y_1|x_1)}{\pi_{\rm ref}(y_1|x_1)}- \beta\log\frac{\pi_r(y_2|x_1)}{\pi_{\rm ref}(y_2|x_1)}\\&-\beta\log\frac{\pi_r(y_1|x_2)}{\pi_{\rm ref}(y_1|x_2)} \\
\Pi_2 = &2\beta\log\frac{\pi_r(y_2|x_2)}{\pi_{\rm ref}(y_2|x_2)}-\beta\log\frac{\pi_r(y_1|x_2)}{\pi_{\rm ref}(y_1|x_2)}\\&-\beta\log\frac{\pi_r(y_2|x_1)}{\pi_{\rm ref}(y_2|x_1)} 
\end{aligned}
\label{bt-adapt-f-sub-eq}
\end{equation}
where $\sigma$ is the sigmoid function.
Therefore, the optimization objective of IOPO is to maximize $p(\mathcal{G}_1 \succ \mathcal{G}_2)$. Motivated by \citet{Rafael-direct-2023-etal}, we can formulate a maximum likelihood loss for a parametrized policy model $\pi_{\theta}$ as follows:

\begin{equation}
\small
\begin{aligned}
\mathcal{L}_{\rm IOPO}(\pi_{\theta}) &= -\mathbb{E}_{i\sim D}\,\bigg\{{\rm log}\bigg[\sigma\bigg(\frac{\Pi_1(\pi_{\theta})+\Pi_2(\pi_{\theta})}{2}\bigg)\bigg]\bigg\}\\
i &= <x_1, y_1, x_2, y_2>
\end{aligned}
\label{bt-adapt-loss-eq}
\end{equation}
\begin{equation}
\small
\begin{aligned}
\Pi_1(\pi_{\theta}) &= \bigg(\underbrace{\beta\log\frac{\pi_{\theta}(y_1|x_1)}{\pi_{\rm ref}(y_1|x_1)}- \beta\log\frac{\pi_{\theta}(y_2|x_1)}{\pi_{\rm ref}(y_2|x_1)}}_{\rm \mathtt{Output}}\bigg)\\&+\bigg(\underbrace{\beta\log\frac{\pi_{\theta}(y_1|x_1)}{\pi_{\rm ref}(y_1|x_1)}-\beta\log\frac{\pi_{\theta}(y_1|x_2)}{\pi_{\rm ref}(y_1|x_2)}}_{\rm \mathtt{Input}}\bigg) \\
\Pi_2(\pi_{\theta}) &= \bigg(\underbrace{\beta\log\frac{\pi_{\theta}(y_2|x_2)}{\pi_{\rm ref}(y_2|x_2)}-\beta\log\frac{\pi_{\theta}(y_1|x_2)}{\pi_{\rm ref}(y_1|x_2)}}_{\rm \mathtt{Output}}\bigg)\\&+\bigg(\underbrace{\beta\log\frac{\pi_{\theta}(y_2|x_2)}{\pi_{\rm ref}(y_2|x_2)}-\beta\log\frac{\pi_{\theta}(y_2|x_1)}{\pi_{\rm ref}(y_2|x_1)}}_{\rm \mathtt{Input}}\bigg) 
\end{aligned}
\label{bt-adapt-f-sub-eq-pi}
\end{equation}
where we mark the preference modeling for $\mathtt{Output}$ and $\mathtt{Input}$ in $\Pi_1(\pi_{\theta})$, $\Pi_2(\pi_{\theta})$.

\begin{table*}[!t]
\tabcolsep=0.09cm
\centering
\small
\begin{tabular}{cccccccccc}
\toprule
\multirow{2}{*}{\textbf{Model}} &\multirow{2}{*}{\textbf{Method}} &\multicolumn{2}{c}{\textbf{\textsc{Trace}}}  &\multicolumn{2}{c}{\textbf{IFEval}} &\multicolumn{3}{c}{\textbf{CFBench}} &{\textbf{\textsc{ComplexBench}}}\\ \cmidrule(r){3-4} \cmidrule(r){5-6} \cmidrule(r){7-9} \cmidrule(r){10-10}
& & $\mathtt{IF}$-$\mathtt{S}$ & $\mathtt{IF}$-$\mathtt{M}$ & $\mathtt{S}$-$\mathtt{Acc}$ & $\mathtt{L}$-$\mathtt{Acc}$ & $\mathtt{CSR}$ & $\mathtt{ISR}$ & $\mathtt{PSR}$ & $\mathtt{DRFR}$ \\
\midrule
\multirow{10}{*}{Qwen2-7B} 
& Instruct &72.5 &54.5	&51.6	&56.4	& 75.8	& 39.1 &50.2 &68.1\\ \cmidrule{2-10}
& SFT & 76.0	&56.1	&52.3	&54.2	&77.8	&40.4	&52.9 &68.2\\ \cmidrule{2-10}
& PPO &77.0 & 57.7	&51.4	&53.8	&76.2	&38.8	&50.6 &68.6 \\ \cmidrule{2-10}
& ORPO &77.9 & 61.7 	& 53.1	& 56.9	& 79.7	& 45.9 & 57.0 & 69.1\\ \cmidrule{2-10}
& SimPO &78.3 & 63.6	& 52.2	& 57.6	& 78.4	& 45.0	& 57.6 & 67.8\\ \cmidrule{2-10}
& DPO &79.0	&67.2	&52.7	&58.2	&80.0	&45.1	&57.9 &70.9 \\ \cmidrule{2-10}
& IOPO (Ours)$_{\textcolor{blue}{Improv.}}$ &82.0$_{\textcolor{blue}{\uparrow 3.0}}$ &68.9$_{\textcolor{blue}{\uparrow 1.7}}$	&59.9$_{\textcolor{blue}{\uparrow 7.2}}$	&63.6$_{\textcolor{blue}{\uparrow 5.4}}$	&80.7$_{\textcolor{blue}{\uparrow 0.7}}$	&47.0$_{\textcolor{blue}{\uparrow 1.9}}$	&58.7$_{\textcolor{blue}{\uparrow 0.8}}$ & 72.6$_{\textcolor{blue}{\uparrow 1.7}}$ \\ \midrule
\multirow{10}{*}{Llama3.1-8B} 
& Instruct &67.5 &52.9 &74.3	&78.6	&71.4	& 35.7	&46.9 & 63.2 \\ \cmidrule{2-10}
& SFT &75.5 &62.9		&71.0	&74.1	&78.4	& 43.2	&54.7 & 68.2 \\ \cmidrule{2-10}
& PPO &75.0 &57.3	&69.9	&72.3	&75.9	&40.9	&50.7 & 68.1 \\ \cmidrule{2-10}
& ORPO & 77.0 & 63.1	& 72.3	& 77.3	& 79.4	& 46.6	&57.2 & 69.6 \\ \cmidrule{2-10}
& SimPO & 76.3 & 64.5	& 71.2	&76.6	& 80.6	& 47.8	& 58.7 & 70.2\\ \cmidrule{2-10}
& DPO &79.0 & 69.2 &71.5	& 76.5	& 80.8	&48.1	&59.8 & 70.8 \\ \cmidrule{2-10}
& IOPO (Ours)$_{\textcolor{blue}{Improv.}}$ & 81.5$_{\textcolor{blue}{\uparrow 2.5}}$ & 70.7$_{\textcolor{blue}{\uparrow 1.5}}$ & 78.2$_{\textcolor{blue}{\uparrow 6.7}}$	& 81.0$_{\textcolor{blue}{\uparrow 4.5}}$	& 81.8$_{\textcolor{blue}{\uparrow 1.0}}$	&49.9$_{\textcolor{blue}{\uparrow 1.8}}$	&61.1$_{\textcolor{blue}{\uparrow 1.3}}$ & 71.8$_{\textcolor{blue}{\uparrow 1.0}}$\\ 
\bottomrule
\end{tabular}
\caption{Main results on in-domain \textsc{Trace}, and out-of-domain IFEval, CFBench, and \textsc{ComplexBench}. $\textcolor{blue}{Improv.}$ indicates the absolute improvement compared to DPO.} %
\label{tb-mainresult}
\end{table*}

\begin{table*}[t]
\centering
\small
\begin{tabular}{cccccccccc}
\toprule
\multirow{2}{*}{\textbf{Model}} &\multirow{2}{*}{\textbf{Method}} &\multicolumn{2}{c}{\textbf{\textsc{Trace}}}  &\multicolumn{2}{c}{\textbf{IFEval}} &\multicolumn{3}{c}{\textbf{CFBench}} &{\textbf{\textsc{ComplexBench}}}\\ \cmidrule(r){3-4} \cmidrule(r){5-6} \cmidrule(r){7-9} \cmidrule(r){10-10}
& & $\mathtt{IF}$-$\mathtt{S}$ & $\mathtt{IF}$-$\mathtt{M}$ & $\mathtt{S}$-$\mathtt{Acc}$ & $\mathtt{L}$-$\mathtt{Acc}$ & $\mathtt{CSR}$ & $\mathtt{ISR}$ & $\mathtt{PSR}$ & $\mathtt{DRFR}$  \\
\midrule
\multirow{4}{*}{Qwen2-7B} 
& IOPO &82.0 &68.9	&59.9	&63.6	&80.7	&47.0	&58.7 & 72.6\\ \cmidrule{2-10}
& w/o $\mathtt{Output}$ $\mathtt{Preference}$ & 81.0 & 66.7	& 55.1	& 60.5	& 79.4	& 46.6	& 56.3 & 71.0 \\ \cmidrule{2-10}
& w/o $\mathtt{Input}$ $\mathtt{Preference}$ & 80.9 & 67.1 & 56.7 & 61.9 & 79.7	& 46.8 &57.0 & 71.3\\ \midrule
\multirow{4}{*}{Llama3.1-8B} 
& IOPO & 81.5 & 70.7 & 78.2	& 81.0	& 81.8	&49.9	&61.1 & 71.8 \\ \cmidrule{2-10}
& w/o $\mathtt{Output}$ $\mathtt{Preference}$ & 81.5 & 69.6 & 77.3 & 80.6 & 80.6 & 48.6 & 58.4 & 69.2 \\ \cmidrule{2-10}
& w/o $\mathtt{Input}$ $\mathtt{Preference}$ & 79.0 & 69.0 & 77.9 & 80.2 & 80.9 & 48.3 & 59.4 & 70.1\\ 
\bottomrule
\end{tabular}
\caption{Ablation studies on \textsc{Trace}, IFEval, CFBench, and \textsc{ComplexBench}.} %
\label{tb-ablation}
\end{table*}

\section{Experiments}
\subsection{Experimental Settings}
\textbf{Evaluation Datasets.}
We conduct experiments on four instruction-following datasets: \textsc{Trace}, IFEval~\cite{zhou2023instruction}, CFBench~\cite{zhang2024cfbench}, and \textsc{ComplexBench}~\cite{wen2024benchmarking}.
\textsc{Trace} evaluation set is introduced in this paper, which has 1,042 instructions, and an average of 4.89 constraints per instruction, with a maximum of 15 constraints.
IFEval consists of 541 prompts, with each prompt containing one or multiple verifiable instructions.
CFBench contains 1,000 samples that cover more than 200 real-life scenarios and over 50 NLP tasks, with each sample including multiple constraints. 
\textsc{ComplexBench} creates 1,150 samples based on 4 constraint types, 19 constraint dimensions, and 4 composition types.
It is worth noting that \textsc{Trace} is the in-domain evaluation set, IFEval, CFBench and \textsc{ComplexBench} are the out-of-domain ones.

\textbf{Implementation Details.}
(1) \textbf{\textsc{Trace} Benchmark}: we choose Qwen2-72B-Instruct~\cite{yang2024qwen2}\footnote{ https://modelscope.cn/models/Qwen/Qwen2-72B-Instruct} for benchmark construction.
(2) \textbf{IOPO Alignment}:
we choose Qwen2-7B-Instruct\footnote{https://modelscope.cn/models/Qwen/Qwen2-7B-Instruct}, and LLaMA3.1-8B-Instruct\footnote{https://huggingface.co/meta-llama/Llama-3.1-8B-Instruct} as the LLM backbone. All models, except for the base models (i.e. Qwen2-7B-Instruct and Llama-3.1-8B-Instruct), are trained on \textsc{Trace}'s training set or its variants, tested on all evaluation datasets ({\it Train Once, Test Anywhere}).
The learning rate is 1e-4 for supervised fine-tuning (SFT), and 5e-6 for DPO and IOPO. The maximum length and epoch are set to 6,000 and 3 respectively. $\beta$ is set to 0.1.
We implement our code based on LLaMA-Factory~\cite{zheng2024llamafactory}, perform parallel training on 4 $\times$ 8-GPU machines, with a micro batch size of 1 per GPU.
The DPO training data construction is shown in Appendix~\ref{dpo-data}.

\textbf{Evaluation Metrics.}
For \textsc{Trace}, we use GPT-4o to evaluate if all constraints in the instruction have been followed (IF-S for single-constraint instructions, and IF-M for multi-constraint instructions), as described in Sec.~\ref{eval_proto}.
For IFEval, we use prompt-level strict and loose accuracy defined in \citet{zhou2023instruction}, abbr. S-Acc and L-Acc respectively.
CFBench~\cite{zhang2024cfbench} introduces three evaluation metrics with GPT-4o as the evaluation model: constraint satisfaction rate (CSR), instruction satisfaction rate (ISR), and priority satisfaction rate (PSR).
\textsc{ComplexBench}~\cite{wen2024benchmarking} calculates the decomposed requirements following ratio (DRFR).

\subsection{Experimental Results}
\textbf{Main Results.}
As shown in Table~\ref{tb-mainresult}, we give the main results under different benchmarks, including in-domain \textsc{Trace}, out-of-domain IFEval, CFBench and \textsc{ComplexBench}. The experiments are conducted under two different base models, Qwen2-7B, and Llama3.1-8B, where Instruct means directly using Qwen2-7B-Instruct or Llama3.1-8B-Instruct for inference, SFT represents the model is trained on \textsc{Trace} training set, and PPO, DPO, IOPO are respectively trained on preference data derived from \textsc{Trace} training set.

For in-domain evaluation on \textsc{Trace} set, we can see 3.0\%, 1.7\% improvements of IOPO on single- and multi-constraint instructions with Qwen2-7B as the base model compared to DPO, and 2.5\%, 1.5\% improvements with Llama3.1-8B as the base model.
For out-of-domain evaluation on IFEval, CFBench and \textsc{ComplexBench}, IOPO achieves an average increase of 2.95\%, and 2.72\% in comparison with DPO based on Qwen2-7B and Llama3.1-8B respectively.
The significant advantages of both in-domain and out-of-domain evaluations confirm the effectiveness of input-output preference optimization, which intensively considers the constraint differences between instructions, enhancing the model's perception of constraints.
It is worth noting that IOPO has a larger performance gap with SFT especially on IFEval, compared to DPO and SFT, which confirms the generalization of IOPO and the necessity of further modeling the input preferences.

\textbf{Ablation Studies.}
To further confirm the effectiveness of input and output preference, we conduct ablation studies on \textsc{Trace}, IFEval, CFBench, and \textsc{ComplexBench} as shown in Table~\ref{tb-ablation}, where ``w/o Output Pref''~\footnote{$$\mathcal{L}_{\text{IOPO}^{*}}(\pi_\theta) = -\mathbb{E}_{i \sim D} { \log [ \sigma ({\Pi^{*}(\pi_\theta)} ) ] }$$
$$i = <x_1, y_1, x_2, y_2>$$ $$\Pi^*(\pi_\theta) = \beta \log \frac{\pi_\theta(y_1|x_1)}{\pi_{\text{ref}}(y_1|x_1)} - \beta \log \frac{\pi_\theta(y_1|x_2)}{\pi_{\text{ref}}(y_1|x_2)}$$ where the preference pair data $y_2|x_2$, $y_2|x_1$ are also used for training.} means we only consider the modeling of input preference with the same training data, ``w/o Input Pref''~\footnote{Different from ``w/o Output Pref'', $$\Pi^*(\pi_\theta) = \beta \log \frac{\pi_\theta(y_1|x_1)}{\pi_{\text{ref}}(y_1|x_1)} - \beta \log \frac{\pi_\theta(y_2|x_1)}{\pi_{\text{ref}}(y_2|x_1)}$$ where the preference pair data $y_2|x_2$, $y_1|x_2$ are also used for training.} means we only consider the modeling of output preference.
We see that {\it output preference} contributes to 2.1\%, and 1.28\% increases with Qwen2-7B and Llama3.1-8B respectively, {\it input preference} separately brings 1.5\% and 1.4\% performance gains, which confirms the effectiveness of both input and output preference modeling.
Besides the paradigm for modeling output preference in existing alignment methods, it's established that modeling input preference is crucial for deeply considering constraints within the instruction.


\begin{table}[t]
\renewcommand\arraystretch{1.1}
\centering
\small
\begin{tabular}{cccc}
\toprule
 \diagbox[width=7em,trim=l]{}{Method}  & SFT & DPO & IOPO\\ \midrule
 \#Memory      & 1$\times$  & 2$\times$ & 4$\times$ \\ 
 \#Training Time     & 14.54 h  & 26.30 h & 34.27 h \\
 \#Inference Speed   & 1$\times$  & 1$\times$ & 1$\times$ \\ 
 \bottomrule
\end{tabular}
\caption{Analysis on the consumed GPU memory, training time, and inference speed under the same batch size.}
\label{tb-param}
\end{table}


\textbf{Complexity Analysis.}
We conduct the analyses of complexity in Table~\ref{tb-param}, where all methods are conducted under the same experimental settings, such as the batch size and GPU. 
(1) For \#Memory, DPO and IOPO are approximately twice and four times that of SFT respectively, because DPO needs a pair of responses to calculate the corresponding loss (<$x, y_1$>, <$x, y_2$>), and IOPO needs to compute four groups of input-output pairs (<$x_1, y_1$>, <$x_2, y_2$>, <$x_1, y_2$>, <$x_2, y_1$>) in its loss.
(2) For \#Training Time, DPO and IOPO require the computation of more tokens compared to SFT under the same batch size, leading to longer training time.
(3) For \#Inference Speed, SFT, DPO, and IOPO are all the same base model optimized for inference, resulting the same inference speed.
The training efficiency and GPU memory usage of IOPO are not the best among compared baselines, but their efficiencies are still of the same order of magnitude, which are reasonable and acceptable comprehensively considering its performance advantage.

\begin{figure}[t]
\centering
\includegraphics[width=1.0\columnwidth]{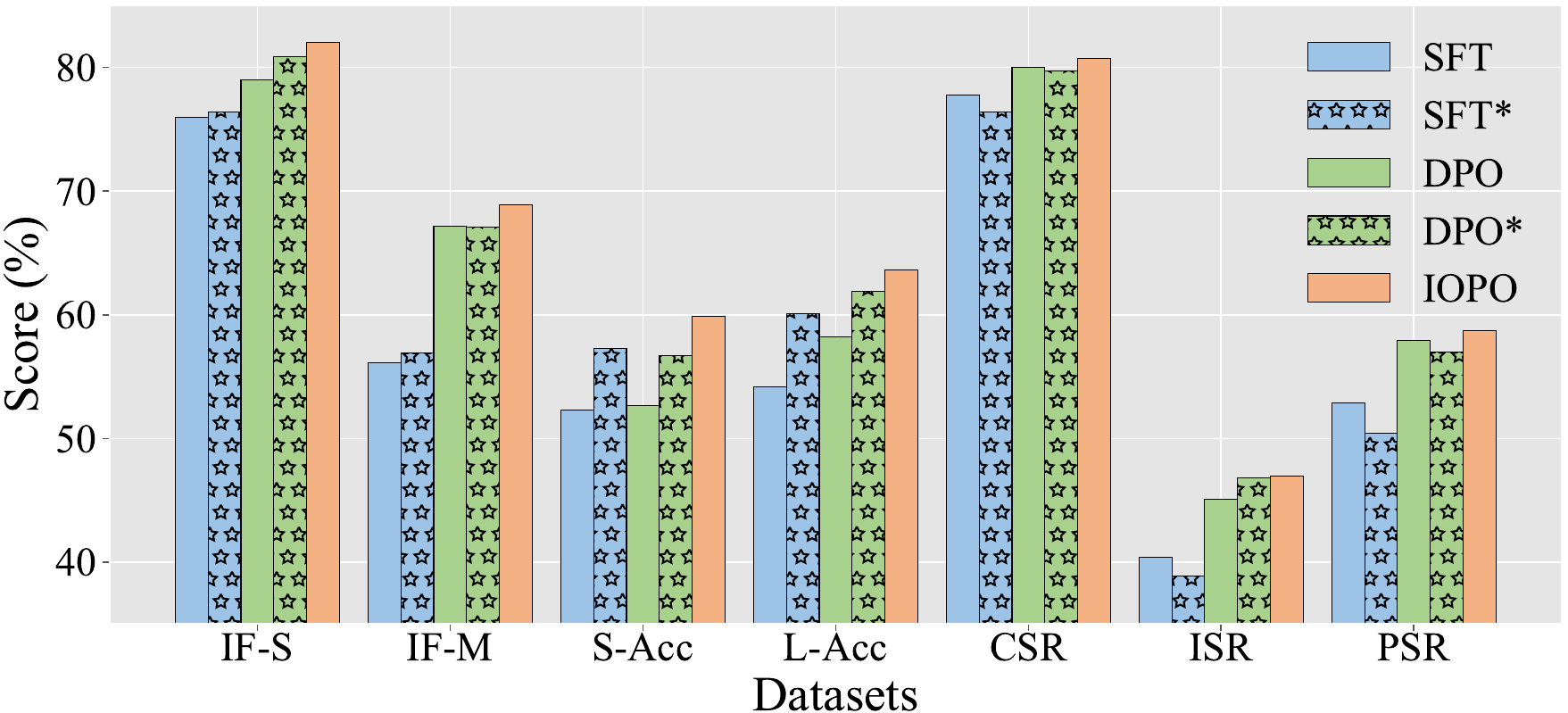} 
\caption{Performance comparisons under the same quantity of tokens with \textbf{Qwen2-7B} as the base model.}
\label{figqwentoken}
\end{figure}

\begin{figure}[t]
\centering
\includegraphics[width=1.0\columnwidth]{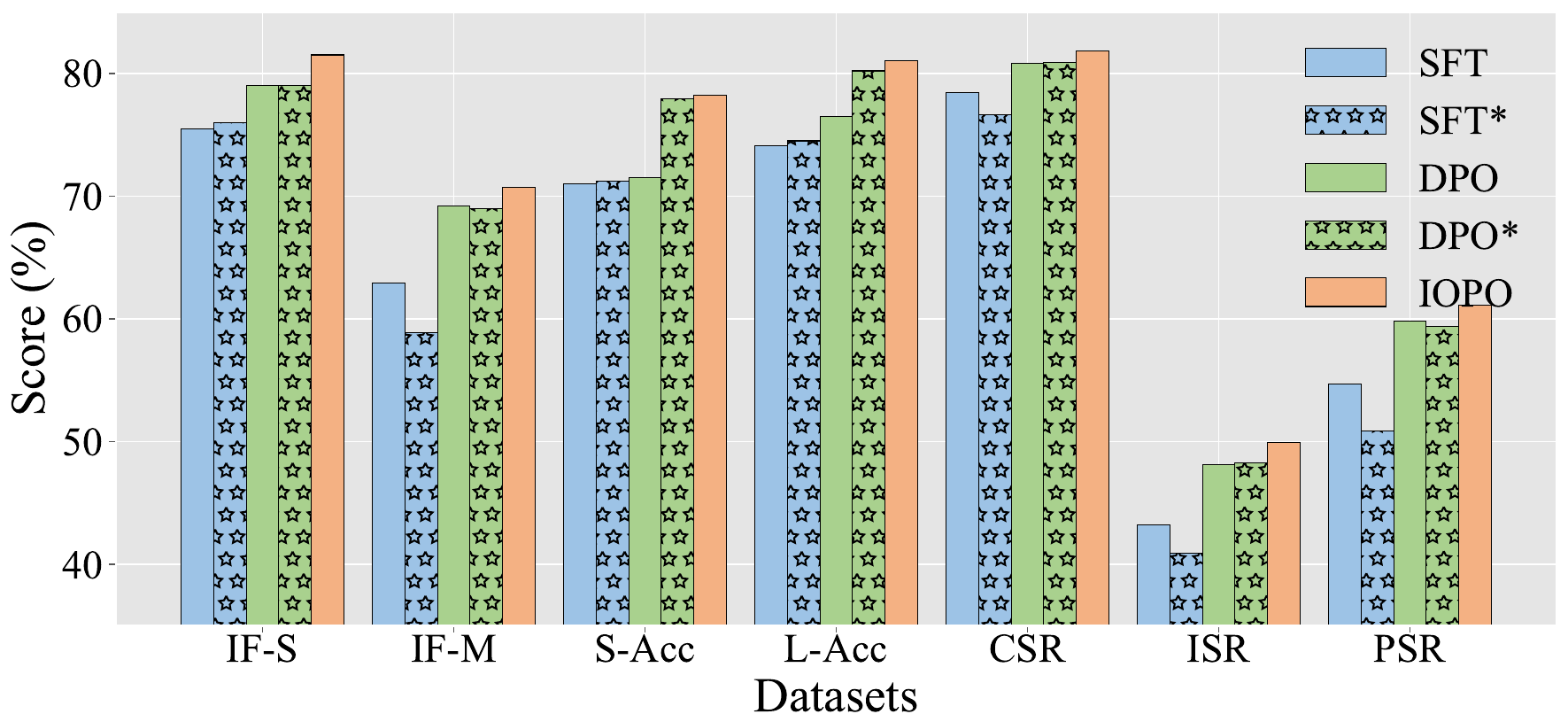} 
\caption{Performance comparisons under the same quantity of tokens with \textbf{Llama3.1-8B} as the base model.}
\label{figllamatoken}
\end{figure}

\textbf{The Impact of Token Quantity.}
To address concerns regarding the IOPO training tokens, we conduct the analyses on the impact of token quantity and report the results in Figure~\ref{figqwentoken}, and Figure~\ref{figllamatoken}.
For IOPO, there exist two instructions along with their corresponding/right responses (\{<$x_1$, $y_1$>, <$x_2$, $y_2$>\}).
To ensure that DPO and IOPO consume the same number of tokens, we construct two pairs of output preferences based on IOPO's instructions $x_1$ ($y_{\rm win}=y_1$, $y_{\rm lose}=y_2$), and $x_2$ ($y_{\rm win}=y_2$, $y_{\rm lose}=y_1$) for training DPO model, denoted by DPO$^*$. Similarly, we train SFT model with instruction data \{<$x_1$, $y_1$>, <$x_2$, $y_2$>\}, denoted by SFT$^*$.
We can observe that increasing the token quantity does indeed yield better performance on some datasets. For example, compared to DPO, DPO$^*$ has achieved a performance improvement on IFEval (S-Acc, L-Acc) with Qwen2-7B as the base model.
However, there are also cases of decline, such as comparing DPO$^*$ with DPO on CFBench (CSR, PSR), which indicates that it is not the case that more tokens always lead to better performance.
At the same time, although consuming the same number of tokens, SFT and DPO still have certain gaps compared to the proposed IOPO, which confirms that the performance improvement of our IOPO does not primarily come from using more tokens, but rather from better constraint-aware modeling of input-output preferences.

\section{Conclusion}
This paper focuses on the ability of LLMs to follow complex instructions, and introduces \textbf{\textsc{Trace}}, a multi-constraint complex instruction benchmark which consists of 120K training samples and 1K test cases.
Furthermore, we propose \textbf{\textsc{IOPO}} alignment method by taking both input and output preferences into account, enabling LLMs to directly learn response preferences and subtly perceive constraints in instructions. The empirical results from extensive testing across in-domain and out-of-domain datasets demonstrate the efficacy of \textsc{IOPO}, with notable improvements of 2.18\% and 3.13\% compared to DPO, respectively. 
For future work, we expect to introduce a more in-depth reasoning process to improve constraint-aware abilities.
\section*{Limitations}
In \textsc{Trace}, the evaluation set has undergone strict manual verification but the training set has not performed this process considering the cost. Although the models trained on the training set have achieved the significant improvements, we believe that if we can further improve the quality of the training set, it will lead to better model performance on effectiveness and generalization of following complex instructions.
\section*{Acknowledgements}
We would like to thank the anonymous reviewers
for their insightful comments and constructive suggestions.
We sincerely thank members of the ConvAI
Team in Tongyi Lab for their valuable feedback and discussions.
\bibliography{ref}
\clearpage
\appendix



\onecolumn
\section{Taxonomy of Constraint}
\label{appendix:tax_constraint}
\scriptsize
\begin{longtable}{ccp{6cm}}
\hline
\textbf{Constraint Type} & \textbf{Constraint Dimension}  & \textbf{Description} \\ \hline
Content Constraint & Theme Constraint  & The generated content should focus on a specific topic or field. \\ \cline{2-3}
& Exclusion Constraint & Clearly specify the information or content that should not be included in the generated content.\\ \cline{2-3}
& Inclusion Constraint & Clearly specify the particular information or content that must be included in the generated content. \\ \cline{2-3}
& Value Constraint & The generated content should not contain information that violates values, such as safety, false information, discrimination, or bias. \\ \cline{2-3}
& Privacy Constraint & The generated content should not include details that may infringe on privacy, such as personal data or sensitive information. \\ \cline{2-3}
& Numerical Constraint & Limit the length and number of words, sentences, and paragraphs in the generated content, or use numerical precision constraints to ensure accuracy. \\
\hline
Situation Constraint & Role-Playing Constraint & The generated content should be based on a specific role or situational background. \\ \cline{2-3}
    & Target Audience Constraint & The generated content should target a specific audience, which affects the terminology used, the level of detail provided, and the complexity of the content. \\ \cline{2-3}
    & Prior Condition Constraint & When a specific intention is met, a particular process should be followed to perform an operation or output specific content. \\ \cline{2-3}
    & \makecell[c]{Natural Language Process\\ Background Information Constraint} & Add natural language form process information, such as procedures or business processes, to assist in generating answers. \\ \cline{2-3}
    & \makecell[c]{Markdown Process\\ Background Information Constraint} & Add markdown-formatted process information, such as procedures or business processes, to assist in generating answers. \\ \cline{2-3}
    & \makecell[c]{Table Background\\ Information Constraint} & Background information is presented in table form, providing a series of markdown-formatted tables to assist in generating answers. \\ \cline{2-3}
    & \makecell[c]{Text Background\\ Information Constraint} & Background information is presented in text form, providing a series of textual background information to assist in generating answers. \\
    \hline
Style Constraint & Tone and Style Constraint & The generated content should adopt a specific tone and style, such as formal, polite, academic, concise, literary, romantic, or sci-fi. \\ \cline{2-3}
  & Emotion Constraint & The generated content should express a specific emotion or mood, such as ensuring the content is positive, inspiring, or empathetic.\\ \cline{2-3}
  & Linguistic Characteristics Constraint & Use specific linguistic features, such as metaphors, personification, and other rhetorical devices. \\ \cline{2-3}
 & Multilingual Constraint & The content should be generated in a specific language or switch between languages according to complex patterns.\\ 
 \hline
 Format Constraint & Output Format Constraint & The generated content should be in a specific data format, such as tables, JSON, HTML, LaTeX, or Markdown. \\ \cline{2-3}
& Text Pattern Constraint & Use specified fonts and font sizes, or special emoji, to ensure readability across different devices and platforms. \\ \cline{2-3}
& Grammar Structure Constraint & The generated content should strictly follow specific grammatical structures, such as subject-predicate-object, subject-verb, etc. \\ \cline{2-3}
& Citation Constraint & The generated content should include citations to sources, providing reliable sources and literature support; follow specific citation formats or reference styles. \\ \cline{2-3}
& Numbering and List Constraint & The generated content should use numbered lists or bullet points to organize information. \\ \cline{2-3}
& Hierarchical Structure Constraint & The generated content should be organized according to a specific hierarchical structure, such as using headings and subheadings. \\ \cline{2-3}
& Template Constraint & The generated content should follow a specific layout or format, such as text alignment, paragraph indentation, and structural templates like introduction-body-conclusion. \\
\hline
Example Constraint & Positive Example Constraint & Provide examples that meet the requirements, and require the model to generate content based on these examples. \\ \cline{2-3}
& Negative Example Constraint & Provide examples that do not meet the requirements, and require the model to avoid generating content similar to these examples. \\
\hline
\caption{Five constraint types and 26 constraint dimensions with their corresponding descriptions.}
\label{tb-constraint-type}\\
\end{longtable}
\normalsize 

\clearpage
\twocolumn
\section{Prompt}
\label{appendix:prompt}

























\subsection{Constraint Expansion Prompt}
\small
\begin{framed}
{{\noindent\textbf{[Task Prompt]}

You are an instruction enhancer. Given an instruction, you need to modify it by adding constraints to make it more complex. You can choose several appropriate types of constraints from those given below, but you must maintain the thematic consistency of the original instruction.

{\it\textbf{\{Constraints\}}}


















\noindent\textbf{[Input]}

—INPUT—

<Instruction>:

{\it \textbf{\{Instruction\}}}

—OUTPUT—}}
\end{framed}
\normalsize

\subsection{Instruction Structuring Prompt}
\small
\begin{framed}
\noindent\textbf{[Task Prompt]}

You are provided with an instruction. As a prompt engineer, your task is to extract the task description, constraints, and the input contained in the given instruction.

\noindent\textbf{[Requirements]}

If there is no constraints information that can be extracted from the instruction, only output NULL in the constraints field.

If there is no input information that can be extracted from the instruction, only output NULL in the input field.

Information in the input field and constraints field cannot be duplicated.

Information in the input field and task description field cannot be duplicated.

Information in the task description field and constraints field cannot be duplicated.

The content extracted for the task description, constraints, and input elements should be consistent with the semantics of the instruction to be extracted.

Evaluate the quality of the instruction; if the instruction is poor, incomplete, or contradictory, do not perform constraints extraction.

\noindent\textbf{[Input]}

—INPUT—

<Instruction>:

{\it\textbf{\{Instruction\}}}

—OUTPUT—
\end{framed}
\normalsize
\subsection{Judge Completeness Prompt}
\small
\begin{framed}
\noindent\textbf{[Task Prompt]}

You are an instruction integrity discriminator, capable of determining whether a given instruction is complete.

\noindent\textbf{[Requirements]}

The given instruction consists of three parts: <Task Description, Constraints, Input>, where Input can be NULL.
You can refer to the examples given in Example, but you should not directly copy the examples.

\noindent\textbf{[Example]}

{\it\textbf{\{Examples\}}}

\noindent\textbf{[Input]}

—INPUT—

{\it\textbf{\{Instruction\}}}

—OUTPUT—
\end{framed}

\subsection{Judge Redundancy Prompt}
\begin{framed}
\noindent\textbf{[Task Prompt]}

You are the redundancy detector for instructions, capable of determining whether given instructions are redundant.

\noindent\textbf{[Requirements]}

The given instruction consists of <Task Description, Constraints, Input>, where Input can be NULL;
You can refer to the Examples provided, but you should not directly copy the examples.

\noindent\textbf{[Example]}

{\it\textbf{\{Examples\}}}






























\noindent\textbf{[Input]}

—INPUT—

{\it\textbf{\{Instruction\}}}

—OUTPUT—
\end{framed}
\normalsize

\subsection{Response Evaluation Prompt}
\small
\begin{framed}
\noindent\textbf{[System]}

You are a fair judge, and please evaluate the quality of an AI assistant’s responses to user query. You need to assess the response based on the following constraints. 
We will provide you with the user's query, some constraints, and the AI assistant's response that needs your evaluation. When you commence your evaluation, you should follow the following process:

1. Evaluate the AI assistant's response on different constraints, and after each constraint evaluation, assign a score from 0 to 10.

2. Aggregate the assessments from each constraint to give an overall score for the AI assistant's response, ranging from 0 to 10.

3. Your scoring should be as strict as possible, overall, the higher the quality of the model's response, the higher the score.

4. When the model's response is irrelevant to the question, or contains significant factual errors, or generates harmful content, the Constraints Overall Score must be 0 points.

5. It is necessary to strictly follow the format in the [Example] for generation, the Fine Grained Score format is Json, and Constraints Overall Score format is List.

Please remember to provide evaluations and explanations before your scoring. After your explanation of each  constraint, include a score for that constraint. 

\noindent\textbf{[Example]}

{\it\textbf{\{Examples\}}}





















\noindent\textbf{[Input]}

—INPUT—

\noindent\#Task Description:

{\it\textbf{\{task\_description\}}}

\noindent\#Constraints:

{\it\textbf{\{constraint\}}}

\noindent\#Input:

{\it\textbf{\{input\}}}

\noindent\#Response:

{\it\textbf{\{answer\}}}

—OUTPUT—
\end{framed}
\normalsize

\section{DPO-Series Data Construction}
\label{dpo-data}

We construct DPO training data based on \textsc{Trace} training set by prompting Qwen2-72B-Instruct to generate a worse response $y_{\rm loose}$ compared to original response $y_{\rm win}$. The construction process is depicted in Figure~\ref{fig_dpo_construct}, the prompt is shown as follows:

\begin{figure}[t]
\centering
\includegraphics[width=1.0\columnwidth]{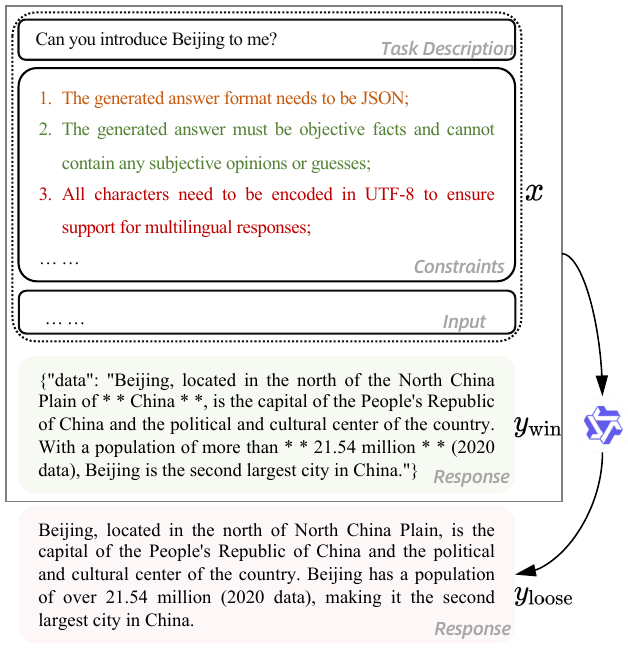} 
\caption{DPO-series Data Construction.}
\label{fig_dpo_construct}
\end{figure}

\small
\begin{framed}
\noindent\#Task Description:

{\it\textbf{\{task\_description\}}}

\noindent\#Constraints:

{\it\textbf{\{constraint\}}}

\noindent\#Input:

{\it\textbf{\{input\}}}

\noindent\#Ref:

The provided answer is: {\it\textbf{\{response\}}}

According to \#Task Description, \#Constraints and \#Input, please generate a \textbf{Worse} answer in terms of complying with the \#Constraint than the provided one.

Please ONLY output the answer.
\end{framed}
\normalsize

\section{IOPO Data Construction}
\label{iopo-data}

\begin{figure*}[t]
\centering
\includegraphics[width=2.0\columnwidth]{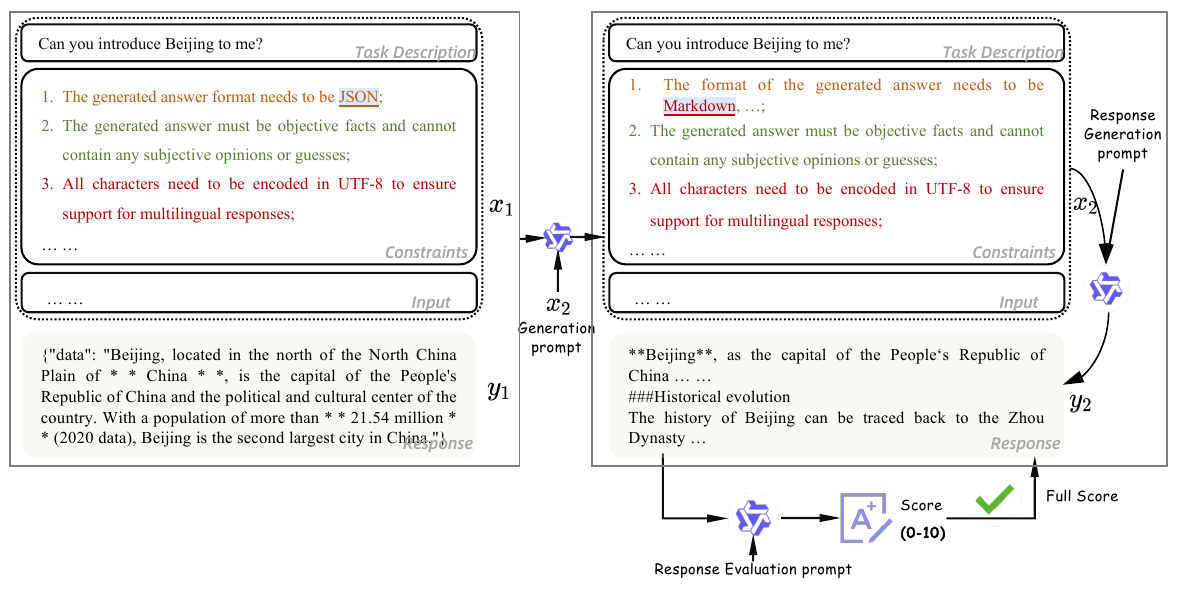} 
\caption{IOPO Data Construction.}
\label{fig_iopo_construct}
\end{figure*}

We construct IOPO training data based on \textsc{Trace} training set by the following steps (the detailed process is shown in Figure~\ref{fig_iopo_construct}):

\textbf{Step 1:} prompting Qwen2-72B-Instruct to generate new constraints by ``add'', ``remove'', and ``revise'' operations, making the response not comply with the new constraints, and then the task description, new constraints, and input are combined to form $x_2$.
The corresponding prompt is as follows:

$x_2$ {\it Generation Prompt}:
\small
\begin{framed}
\noindent\#Task Description:

{\it\textbf{\{task\_description\}}}

\noindent\#Constraints:

{\it\textbf{\{constraint\}}}

\noindent\#Input:

{\it\textbf{\{input\}}}

\noindent\#Ref:

The provided answer is: {\it\textbf{\{response\}}}

According to \#Task Description, \#Constraints and \#Input, please {\it\textbf{\{OP\}}} items of original CONSTRAINTS to generate the new CONSTRAINTS, making the provided answer NOT comply with the new CONSTRAINTS.

Please ONLY output the new CONSTRAINTS.
\end{framed}
\normalsize
{\it\textbf{OP}} can be randomly selected from \{``ADD new items into the'', ``DELETE partial'', ``REVISE specific''\} according to a uniform distribution.

\textbf{Step 2:}
For instruction $x_2$, we prompt Qwen2-72B-Instruct to generate the corresponding response $y_2$. The prompt is {\it Response Generation Prompt}.

\textbf{Step 3:}
We finally prompt Qwen2-72B-Instruct to evaluate the response $y_2$, and only keep the full-score ones. The prompt is {\it Response Evaluation Prompt}.

Finally, we prompt Qwen2-72B-Instruct to check the rationality of the group pairs (<$x_1, y_1$>, <$x_2, y_2$>, <$x_1, y_2$>, <$x_2, y_1$>).

\begin{table*}[!t]
\renewcommand\arraystretch{1.2}
\tabcolsep=0.09cm
\centering
\small
\begin{tabular}{cccccccccc}
\toprule
\multirow{2}{*}{\textbf{Model}} &\multirow{2}{*}{\textbf{Method}} &\multicolumn{2}{c}{\textbf{\textsc{Trace}}}  &\multicolumn{2}{c}{\textbf{IFEval}} &\multicolumn{3}{c}{\textbf{CFBench}} &{\textbf{\textsc{ComplexBench}}}\\ \cmidrule(r){3-4} \cmidrule(r){5-6} \cmidrule(r){7-9} \cmidrule(r){10-10}
& & $\mathtt{IF}$-$\mathtt{S}$ & $\mathtt{IF}$-$\mathtt{M}$ & $\mathtt{S}$-$\mathtt{Acc}$ & $\mathtt{L}$-$\mathtt{Acc}$ & $\mathtt{CSR}$ & $\mathtt{ISR}$ & $\mathtt{PSR}$ & $\mathtt{DRFR}$ \\
\midrule
\multirow{10}{*}{Qwen2-7B} 
& Instruct &72.5 &54.5	&51.6	&56.4	& 75.8	& 39.1 &50.2 &68.1\\ \cmidrule{2-10}
& SFT & 76.0	&56.1	&52.3	&54.2	&77.8	&40.4	&52.9 &68.2\\ \cmidrule{2-10}
& PPO &77.0 & 57.7	&51.4	&53.8	&76.2	&38.8	&50.6 &68.6 \\ \cmidrule{2-10}
& ORPO &77.9 & 61.7 	& 53.1	& 56.9	& 79.7	& 45.9 & 57.0 & 69.1\\ \cmidrule{2-10}
& SimPO &78.3 & 63.6	& 52.2	& 57.6	& 78.4	& 45.0	& 57.6 & 67.8\\ \cmidrule{2-10}
& DPO &79.0	&67.2	&52.7	&58.2	&80.0	&45.1	&57.9 &70.9 \\ \cmidrule{2-10}
& IOPO (Ours)$_{\textcolor{blue}{Improv.}}$ &82.0$_{\textcolor{blue}{\uparrow 3.0}}$ &68.9$_{\textcolor{blue}{\uparrow 1.7}}$	&59.9$_{\textcolor{blue}{\uparrow 7.2}}$	&63.6$_{\textcolor{blue}{\uparrow 5.4}}$	&80.7$_{\textcolor{blue}{\uparrow 0.7}}$	&47.0$_{\textcolor{blue}{\uparrow 1.9}}$	&58.7$_{\textcolor{blue}{\uparrow 0.8}}$ & 72.6$_{\textcolor{blue}{\uparrow 1.7}}$ \\ \midrule
\multirow{10}{*}{Llama3.1-8B} 
& Instruct &67.5 &52.9 &74.3	&78.6	&71.4	& 35.7	&46.9 & 63.2 \\ \cmidrule{2-10}
& SFT &75.5 &62.9		&71.0	&74.1	&78.4	& 43.2	&54.7 & 68.2 \\ \cmidrule{2-10}
& PPO &75.0 &57.3	&69.9	&72.3	&75.9	&40.9	&50.7 & 68.1 \\ \cmidrule{2-10}
& ORPO & 77.0 & 63.1	& 72.3	& 77.3	& 79.4	& 46.6	&57.2 & 69.6 \\ \cmidrule{2-10}
& SimPO & 76.3 & 64.5	& 71.2	&76.6	& 80.6	& 47.8	& 58.7 & 70.2\\ \cmidrule{2-10}
& DPO &79.0 & 69.2 &71.5	& 76.5	& 80.8	&48.1	&59.8 & 70.8 \\ \cmidrule{2-10}
& IOPO (Ours)$_{\textcolor{blue}{Improv.}}$ & 81.5$_{\textcolor{blue}{\uparrow 2.5}}$ & 70.7$_{\textcolor{blue}{\uparrow 1.5}}$ & 78.2$_{\textcolor{blue}{\uparrow 6.7}}$	& 81.0$_{\textcolor{blue}{\uparrow 4.5}}$	& 81.8$_{\textcolor{blue}{\uparrow 1.0}}$	&49.9$_{\textcolor{blue}{\uparrow 1.8}}$	&61.1$_{\textcolor{blue}{\uparrow 1.3}}$ & 71.8$_{\textcolor{blue}{\uparrow 1.0}}$\\ \hdashline
\multirow{5}{*}{\makecell[c]{Mainstream\\ LLMs}} 
& OpenAI o1 &85.2 &74.7 &84.6	&89.7	&86.7	& 62.4	&72.1 & 81.3 \\ \cmidrule{2-10}
& DeepSeek-R1 (671B) &84.1 &73.8		&83.5	&89.4	&86.5	& 62.1	&71.9 & 78.7 \\ \cmidrule{2-10}
& Claude-3.5-Sonnet &87.9 &76.3		&86.7	&90.5	&87.1	& 62.6	&72.3 & 82.6 \\
\bottomrule
\end{tabular}
\caption{Performance on \textsc{Trace}, IFEval, CFBench, and \textsc{ComplexBench}.} %
\label{tb-mainresult-appendix}
\end{table*}

\section{IOPO Variant}
In fact, when implementing IOPO in the business applications, we can adopt the following simplification by omitting $\Pi_2(\pi_\theta)$ in Eq.\ref{bt-adapt-loss-eq} for less memory usage:
\begin{equation}
\small
\begin{aligned}
\mathcal{L}_{\rm IOPO^*}(\pi_{\theta}) &= -\mathbb{E}_{i\sim D}\,\bigg\{{\rm log}\bigg[\sigma\bigg(2\beta\log\frac{\pi_{\theta}(y_1|x_1)}{\pi_{\rm ref}(y_1|x_1)}\\&- \beta\log\frac{\pi_{\theta}(y_2|x_1)}{\pi_{\rm ref}(y_2|x_1)}-\beta\log\frac{\pi_{\theta}(y_1|x_2)}{\pi_{\rm ref}(y_1|x_2)}\bigg)\bigg]\bigg\}\\
i &= <x_1, y_1, x_2, y_2>
\end{aligned}
\label{bt-adapt-loss-eq-variant}
\end{equation}
After simplification, the memory usage is approximately 1.5 times that of DPO, and the performance achieves average increase of 2.03\% compared to DPO, decreases by 0.48\% compared to before IOPO simplification.

\section{Compared to Mainstream LLMs}
As shown in Table~\ref{tb-mainresult-appendix}, we evaluate some mainstream closed-sourced/SOTA models, including OpenAI o1~\cite{jaech2024openai}, DeepSeek-R1~\cite{guo2025deepseek}, and Claude-3.5-Sonnet~\cite{anthropic2024claude}, and observe that: (1) Our benchmark \textsc{Trace} provides the consistent rank (Claude-3.5-Sonnet > o1 > DeepSeek-R1) as existing benchmarks IFEval, CFBench, and \textsc{ComplexBench}, confirming the data's quality.
(2) Larger closed-sourced/SOTA models have better performance than 7B/8B weaker models. However, the proposed algorithm IOPO significantly narrows the gap between closed-sourced/SOTA models and 7B/8B base weaker models, especially on in-domain \textsc{Trace} data. This inspires us to construct more diverse high-quality data and more advanced algorithm for enabling weaker models to achieve similar performance to larger-scale models.

\section{Derivation for $p(\mathcal{G}_1 \succ \mathcal{G}_2)$}
\label{appendix:derivation}

As described in Eq.~\ref{bt-adapt-eq} as follows:
\begin{equation}
\small
\begin{aligned}
p(\mathcal{G}_1 \succ \mathcal{G}_2) = \frac{e^{r(x_1, y_1)+r(x_2, y_2)}}{e^{r(x_1, y_1)+r(x_2, y_2)} + e^{r(x_1, y_2)+r(x_2, y_1)}}
\label{bt-adapt-eq-appendix}
\end{aligned}
\end{equation}

As described in Eq.~\ref{reward-eq}, the reward function $r(x, y)$ can be represented by the policy model $\pi_r$ as follows:
\small
\begin{align}
   r(x, y) = \beta \log \frac{\pi_r(y|x)}{\pi_{\text{ref}}(y|x)} + \beta \log Z(x)
\label{reward-eq-appendix}
\end{align}
\normalsize

Combining above equations, we can derive that:

\begin{center}
\small
\rotatebox{270}{%
\begin{minipage}{1.3\textwidth}
\begin{equation}
\begin{aligned}
p^* 
&= \frac{
    \exp\left(\beta\log\frac{\pi^*(y_1|x_1)}{\pi_{\rm ref}(y_1|x_1)}+\beta\log Z(x_1)+\beta\log\frac{\pi^*(y_2|x_2)}{\pi_{\rm ref}(y_2|x_2)}+\beta\log Z(x_2)\right)
}{
    \exp\left(\beta\log\frac{\pi^*(y_1|x_1)}{\pi_{\rm ref}(y_1|x_1)}+\beta\log Z(x_1)+\beta\log\frac{\pi^*(y_2|x_2)}{\pi_{\rm ref}(y_2|x_2)}+\beta\log Z(x_2)\right)+\exp\left(\beta\log\frac{\pi^*(y_2|x_1)}{\pi_{\rm ref}(y_2|x_1)}+\beta\log Z(x_1)+\beta\log\frac{\pi^*(y_1|x_2)}{\pi_{\rm ref}(y_1|x_2)}+\beta\log Z(x_2)\right)
}\\
&= \frac{1}{1+\exp\left(\beta\log\frac{\pi^*(y_2|x_1)}{\pi_{\rm ref}(y_2|x_1)}-\beta\log\frac{\pi^*(y_1|x_1)}{\pi_{\rm ref}(y_1|x_1)}+\beta\log\frac{\pi^*(y_1|x_2)}{\pi_{\rm ref}(y_1|x_2)}-\beta\log\frac{\pi^*(y_2|x_2)}{\pi_{\rm ref}(y_2|x_2)}\right)}\\
&= \sigma\left(\frac{1}{2}(\underbrace{2\beta\log\frac{\pi^*(y_1|x_1)}{\pi_{\rm ref}(y_1|x_1)}-\beta\log\frac{\pi^*(y_2|x_1)}{\pi_{\rm ref}(y_2|x_1)}-\beta\log\frac{\pi^*(y_1|x_2)}{\pi_{\rm ref}(y_1|x_2)}}_{<x_1,y_1>\,{\rm for}\,{\rm different}\,x, y}+\underbrace{2\beta\log\frac{\pi^*(y_2|x_2)}{\pi_{\rm ref}(y_2|x_2)}-\beta\log\frac{\pi^*(y_1|x_2)}{\pi_{\rm ref}(y_1|x_2)}-\beta\log\frac{\pi^*(y_2|x_1)}{\pi_{\rm ref}(y_2|x_1)}}_{<x_2,y_2>\,{\rm for}\,{\rm different}\,x, y})\right)
\end{aligned}
\end{equation}
\end{minipage}}
\end{center}
where the notations $p^*$ and $\pi^*$ are, respectively, equivalent to $p(\cdot)$ and $\pi_{\theta}$ in the main text.



\end{document}